\documentclass{article}

\usepackage[preprint]{neurips_2026}
\usepackage[utf8]{inputenc} 
\usepackage[T1]{fontenc}    
\usepackage{hyperref}       
\usepackage{url}            
\usepackage{booktabs}       
\usepackage{amsmath,amsfonts,amssymb}
\usepackage{nicefrac}       
\usepackage{microtype}      
\usepackage{graphicx}
\usepackage{xcolor}         
\usepackage{float}          
\usepackage{enumitem}
\graphicspath{{figures/}}

\definecolor{withinblue}{HTML}{4C78A8}
\definecolor{crosspurple}{HTML}{7357A6}
\newcommand{\weakreact}{weak-ReAct}
\newcommand{\pp}{\,\mathrm{pp}}
\newcommand{\Within}{\ifmmode\mathit{Within}\else\textit{Within}\fi}
\newcommand{\Cross}{\ifmmode\mathit{Cross}\else\textit{Cross}\fi}
 \newcommand{\PfNevals}{24{,}000}
\newcommand{\PfCoverage}{99.1}
\newcommand{\PfColMin}{2.14}
\newcommand{\PfColMax}{9.27}
\newcommand{\PfRowMin}{5.55}
\newcommand{\PfRowMax}{6.46}

\newcommand{\PfSeenGain}{+0.77}

\newcommand{\PfSeenCIcorr}{$[-0.21, +1.75]$}

\newcommand{\PfHeldKeight}{+0.13}

\newcommand{\PfCrossWithinKeight}{+0.25}
\newcommand{\PfCrossWithinKeightCI}{$[-0.48, +1.02]$}
\newcommand{\PfWithinSftKeight}{-0.03}

\newcommand{\PfCrossWithinKfour}{+0.58}
\newcommand{\PfSeedCW}{+0.16}
\newcommand{\PfSeedCWCI}{$[-0.41, +0.72]$}
\newcommand{\PfSeedRangeW}{0.45}
\newcommand{\PfSeedRangeC}{0.42}
\newcommand{\PfAdvWithin}{+0.02}

\newcommand{\PfAdvCross}{+4.48}
\newcommand{\PfAdvCrossCI}{$[+3.22, +5.83]$}
\newcommand{\PfAdvCov}{+1.17}

\newcommand{\PfAdvResid}{+4.54}

\newcommand{\PfMdeKfour}{1.19--1.42}
\newcommand{\PfMdeKeight}{0.88--1.14}
\newcommand{\PfQsix}{-0.03}
\newcommand{\PfQsixCI}{$[-1.03, +0.95]$}
\newcommand{\PfOnlineCross}{+0.13}
\newcommand{\PfOnlineCrossCI}{$[-0.85, +1.10]$}
\newcommand{\PfOnlineWithin}{+0.75}
\newcommand{\PfOnlineWithinCI}{$[-0.10, +1.65]$}

\newcommand{\PfJsdWithinMin}{0.0003}
\newcommand{\PfJsdWithinMax}{0.0029}
\newcommand{\PfJsdAcrossMin}{0.17}
\newcommand{\PfJsdAcrossMax}{0.35}
\newcommand{\PfJsdRatio}{59}
\newcommand{\PfActWrite}{-1.35}
\newcommand{\PfActWriteCI}{$[-2.30, -0.40]$}
\newcommand{\PfActRun}{+0.66}
\newcommand{\PfActRunCI}{$[+0.29, +1.01]$}
\newcommand{\PfActVcs}{+1.28}
\newcommand{\PfActVcsCI}{$[+0.24, +2.40]$}

\newenvironment{paperfigure}[1][tbp]
  {\begin{figure}[#1]}{\end{figure}}
\newenvironment{papertable}[1][tbp]
  {\begin{table}[#1]}{\end{table}}
\newcommand{\papertablecaption}[2]{\caption{#2}\label{#1}}
\newcommand{\paperlayout}[2]{#1}

\hypersetup{colorlinks=true, linkcolor=blue!60!black, citecolor=blue!60!black, urlcolor=blue!60!black}

\title{Multi-Harness Reinforcement Learning for Coding Agents:\paperlayout{}{\\}
Training Gains and Harness Dependence}

\author{%
  Chenqian Le\thanks{Equal contribution.} \\
  New York University \\
  \texttt{cl6707@nyu.edu}
  \And
  Jiayi Cheng\footnotemark[1] \\
  New York University \\
  \texttt{jc10077@nyu.edu}
  \paperlayout{\And}{\AND}
  Qijia He \\
  University of Washington \\
  \texttt{heqj3@uw.edu}
  \And
  Runhao Li \\
  University of Southern California \\
  \texttt{runhaoli@usc.edu}
  \paperlayout{\And}{\AND}
  Yinghao Li \\
  Columbia University \\
  \texttt{yinghao.li@columbia.edu}
  \And
  Xupeng Chen\thanks{Corresponding author.} \\
  Dimension Gate \\
  \texttt{xupengchen@dimensiongate.cn}
}

\begin{document}
\raggedbottom
\maketitle
\begin{abstract}
Recent systems train coding agents with reinforcement learning (RL) on trajectories collected
through multiple execution harnesses. However, gains under the training harnesses leave open
whether the improvements transfer to unseen interfaces and how they depend on reward
normalization. We present a controlled
study that measures RL gains under fixed harnesses and compares the same models across harnesses.
We train two Qwen3-8B RL variants from the same supervised fine-tuning (SFT) checkpoint using
identical replay data. They differ only in whether rewards for each task are normalized within
each harness (\Within{}) or across harnesses (\Cross{}).
On SWE-bench Verified, we identify three main findings:
(i)~Multi-harness RL yields modest gains over the SFT baseline under
both normalization rules. Across the four training harnesses, the estimated mean improvements
are $+0.48$ and $\PfSeenGain$ percentage points for \Within{} and \Cross{}, respectively.
(ii)~\Cross{} increases the fraction of trajectories with nonzero advantage but provides no
consistent advantage over \Within{} on an unseen harness across training seeds and evaluation
budgets; neither rule significantly improves over SFT on this harness.
(iii)~Performance varies substantially more across harnesses than with RL training. After
averaging over model variants, the highest and lowest source-harness solve rates differ by
$7.13$ percentage points, far exceeding the estimated training gains.
Even after multi-harness RL, the observed performance differences across execution interfaces
remain substantially larger than either the gains from training or the differences between
normalization rules.
\end{abstract}

\section{Introduction}
\label{sec:intro}

Coding agents interact with repositories through execution harnesses that provide tools for
inspecting files, editing code, and running tests \citep{yang2024sweagent,wang2025openhands}.
Recent systems train a shared policy with reinforcement learning (RL) on trajectories collected
through multiple harnesses or harness configurations
\citep{openforge2026,kimi2026k3,clawgymii2026}. A policy may perform better on the training
harnesses partly because it learns to use their tools and feedback more effectively.
Interpreting these gains therefore requires distinguishing improvements under the training
harnesses from improvements that transfer to other execution interfaces.

Learning from multiple harnesses and comparing rewards across harnesses are two distinct
training choices. For a given task, group-relative policy optimization (GRPO)
\citep{shao2024deepseekmath} can use the same trajectories while normalizing rewards separately
within each harness or jointly across harnesses. Suppose all attempts at a task succeed under
harness A and fail under harness B. With binary success rewards, within-harness normalization
assigns zero advantage to every attempt \citep{yu2025dapo}. Pooling encourages the actions in A's
successful attempts and discourages those in B's failed attempts. Pooling thus adds a learning
signal without adding training data. The successful attempts were made using A's tools and
feedback, so learning from them may improve behavior specific to A. Comparing these rules on
the same trajectories tests whether reward grouping affects RL gains over a supervised
fine-tuning (SFT) baseline under the training harnesses and an unseen harness.

In GRPO, reward normalization and group construction determine
the advantages assigned to sampled trajectories. Dr.~GRPO analyzes how reward normalization
changes the relative contribution of different questions to policy updates
\citep{liu2025drgrpo}, while GiGPO groups actions taken from the same
environment state to assign step-level credit \citep{feng2025gigpo}. Multi-harness systems
introduce another grouping choice: HarnessX pools same-task traces across successive harness
and model versions \citep{harnessx2026}, whereas ClawGym II normalizes rewards within each
task--harness pair \citep{clawgymii2026}. These multi-harness systems also differ in the
trajectories they collect and how they train, so their overall results do not isolate the
effect of cross-harness normalization. Isolating the effect of reward grouping
requires holding the training data and optimization budget fixed.

We conduct a controlled study of multi-harness RL in repository-level coding
(Figure~\ref{fig:design}). Starting
from a common Qwen3-8B \citep{yang2025qwen3} supervised checkpoint, two RL arms replay the same
frozen trajectories collected on SWE-Gym \citep{pan2025swegym} using four execution harnesses:
Aider \citep{gauthier2024aider}, OpenHands \citep{wang2025openhands}, Qwen Code
\citep{qwencode2025}, and SWE-agent \citep{yang2024sweagent}. The arms share the reward
definition and update budget, and differ only in
how rewards are normalized: \Within{} normalizes separately for each task--harness pair,
while \Cross{} pools harnesses within each task. We evaluate both RL policies and the SFT
baseline on SWE-bench Verified under the four source harnesses and a minimal harness excluded
from training. The same hidden
tests score solutions under all five harnesses, allowing us to measure performance on the
training interfaces and transfer to the unseen one.

\begin{paperfigure}[t]
  \centering
  \includegraphics[width=\linewidth]{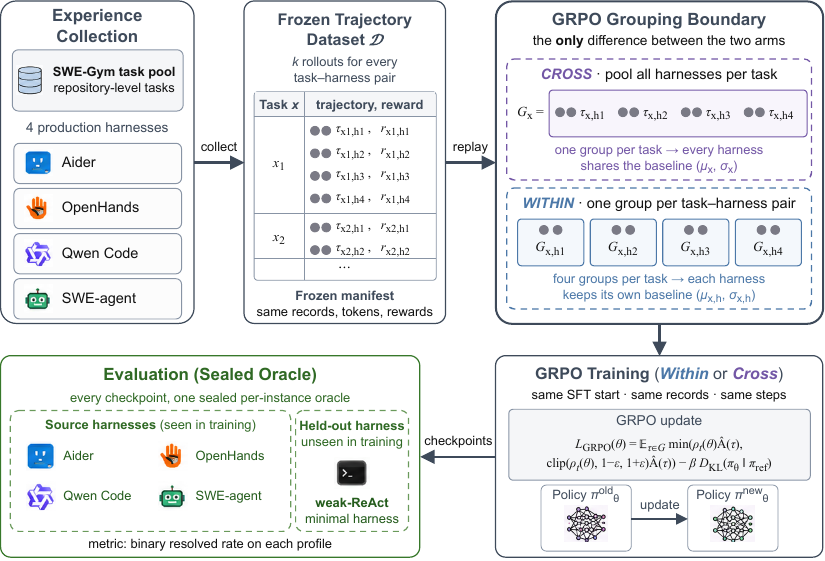}
  \caption{Controlled study of multi-harness RL. Four harnesses
  collect trajectories on SWE-Gym. \Within{} and \Cross{} replay the same frozen dataset from a common
  SFT checkpoint, with identical rewards and update budgets. \Within{} normalizes rewards separately
  for each task--harness pair; \Cross{} normalizes them jointly across harnesses for each task. Both
  RL policies and the SFT baseline are evaluated on SWE-bench Verified using the same hidden tests
  under the four source harnesses and an unseen \weakreact{} harness. Comparisons with SFT measure
  training gains and their transfer to the unseen harness; comparing \Cross{} with \Within{}
  isolates the effect of reward grouping. Evaluating each policy across harnesses measures
  performance variation associated with the execution interface.}
  \label{fig:design}
  \vspace{-0.8mm}
\end{paperfigure}

In our controlled study, multi-harness RL yields modest estimated gains over SFT under both
normalization rules on the source harnesses, while performance differences across evaluation
harnesses are substantially larger. Averaged over the four source harnesses, the estimated
gains are $+0.48$ and $\PfSeenGain$ percentage points for \Within{} and \Cross{}, respectively.
On the held-out harness, neither rule significantly improves over SFT, and \Cross{} provides no
consistent advantage over \Within{} across training seeds and evaluation budgets. The mean
\Cross{}--\Within{} difference over three training seeds is \PfSeedCW{}
percentage points, with a $95\%$ confidence interval of \PfSeedCWCI{}.
The variation across source harnesses is substantially larger than the observed training
gains: after averaging over model variants, the highest and lowest harness solve rates differ
by $7.13$ percentage points. Under a fixed harness, the SFT and \Cross{}
policies exhibit similar frequencies of search, editing, testing, and other actions. Evaluating
the same policy under a different harness produces substantially larger changes in these
frequencies.

We contribute a controlled empirical study of multi-harness RL that quantifies gains over
SFT and compares their magnitude with performance differences across evaluation harnesses.
Holding the training dataset and optimization settings fixed, we examine how reward grouping
affects these gains and evaluate transfer to an unseen harness across training seeds and
evaluation budgets. Advantage diagnostics characterize how reward grouping changes the
fraction of trajectories with nonzero advantages and the association between advantage
estimates and harness identity. Across the tested configurations, the estimated source-harness
gains from multi-harness RL and the differences between normalization rules are both small
relative to performance variation across evaluation harnesses.

\section{Related work}
\label{sec:related}

\paragraph{Reinforcement Learning for LLM Agents.}
Reinforcement learning trains LLM agents through interaction with external environments
\citep{wang2025ragen}, including device control \citep{bai2024digirl}, web navigation
\citep{qi2025webrl}, and search \citep{jin2025searchr1}.
In software engineering, SWE-RL learns from software evolution data using a code-similarity
reward \citep{wei2025swerl}, while SWE-Gym provides executable tasks for trajectory fine-tuning
and verifier training \citep{pan2025swegym}. Polar and OpenForgeRL provide infrastructure for
RL through existing harnesses \citep{polar2026,openforge2026}, and Tmax develops an
outcome-reward training recipe for terminal agents \citep{tmax2026}.
Training across multiple harnesses introduces choices about both rollout allocation and
advantage estimation. Kimi K3 varies harness configurations during training \citep{kimi2026k3},
and HarnessBandit selects a harness for each training batch using advantage magnitude and
gradient alignment across harnesses \citep{wei2026harnessbandit}. For advantage estimation,
HarnessX pools same-task trajectories across successive harness and model versions
\citep{harnessx2026}, whereas ClawGym II normalizes rewards within task--harness pairs
\citep{clawgymii2026}. Our fixed-data study evaluates RL gains across source and held-out
interfaces and compares these grouping rules under the same optimization settings.

\paragraph{Reward Design and Advantage Estimation.}
Outcome and process supervision differ in whether feedback is assigned to complete solutions
or intermediate reasoning steps \citep{uesato2022process,lightman2024letsverify,wang2024mathshepherd}.
Given a reward function, GRPO and RLOO estimate advantages from groups of responses to the same
prompt without a learned value function \citep{shao2024deepseekmath,ahmadian2024rloo}.
Dr.~GRPO removes normalization by response length and reward standard deviation to address
length and difficulty biases \citep{liu2025drgrpo}, while DAPO modifies token weighting and
uses dynamic sampling to exclude groups with identical rewards \citep{yu2025dapo}. Group construction also
supports temporal credit assignment: GiGPO groups actions taken at the same environment state
to estimate step-level advantages in multi-turn trajectories \citep{feng2025gigpo}.
Our study examines the grouping of complete trajectories collected under different harnesses.
We keep the binary terminal reward fixed and vary the group used to compute its mean and
standard deviation: \Within{} uses each task--harness pair, and \Cross{} pools trajectories
across harnesses for each task. This comparison tests how the normalization group affects
advantage estimates and subsequent policy generalization.

\paragraph{Evaluation and Generalization of LLM Agents.}
Agent benchmarks specify tasks, interaction environments, and tool interfaces for evaluating
policies \citep{liu2024agentbench,zhou2024webarena,yao2024taubench}. Within these settings,
measured performance also depends on the execution configuration. Harness-Bench documents
variation in success rates, efficiency, and failure modes across model--harness pairings
\citep{yao2026harnessbench}, while evaluation studies emphasize harness disclosure, cost
measurement, and reproducibility \citep{zhang2026stopcomparing,halharness2025,kapoor2024agents}.
For inference-time recovery, ARCHER uses execution feedback to select among reflection,
replanning, and escalation to a stronger model, with conformal calibration to control expected
recovery cost \citep{he2026archer}.
Studies of RL post-training further examine generalization under shifts in input distributions,
task rules, and visual observations \citep{kirk2024rlhf,chu2025sft}. Our study focuses on a
shift in the execution harness, including its tools and feedback. Evaluating the same policies
under source and held-out harnesses places training gains in the context of interface
differences and tests whether those gains persist under an unseen harness.

\section{Study Design}
\label{sec:setup}

We measure training gains from multi-harness trajectories and compare their magnitude across
execution interfaces (Figure~\ref{fig:design}). Two RL variants (\Within{} and \Cross{}) share
the training data, initialization, and optimization settings, differing only in reward normalization.
Comparisons with SFT measure training gains, while evaluations under source and unseen
harnesses characterize dependence on the execution interface and transfer beyond training.

\subsection{Policy Optimization and Advantage Estimation}
\label{sec:creditrules}

Advantage estimates weight the reward-dependent term in policy optimization. For a trajectory
$\tau$ with advantage $A_\tau$, let $a_t$ denote a model-generated token and $s_t$ its preceding
interaction history. Standard GRPO maximizes an objective built from per-token terms of the form
\citep{shao2024deepseekmath}:
\begin{equation}
\begin{aligned}
j_t(\theta)
={}&\min\Biggl\{
  \frac{\pi_\theta(a_t\mid s_t)}{\pi_{\mathrm{old}}(a_t\mid s_t)}A_\tau,\\
&\operatorname{clip}\!\left(
  \frac{\pi_\theta(a_t\mid s_t)}{\pi_{\mathrm{old}}(a_t\mid s_t)},
  1-\delta,1+\delta\right)A_\tau\Biggr\}\\
&-\beta D_{\mathrm{KL}}\!\left[
  \pi_\theta(\cdot\mid s_t)\,\Vert\,\pi_{\mathrm{ref}}(\cdot\mid s_t)
  \right].
\end{aligned}
\label{eq:policy_objective}
\end{equation}
Here $\pi_{\mathrm{old}}$ is the policy that generated the trajectory,
$\pi_{\mathrm{ref}}$ is the reference policy, $\delta$ is the clipping threshold, and $\beta$
is the KL coefficient. For the reward-dependent term, positive advantages favor higher
probabilities of the sampled tokens, while negative advantages favor lower probabilities,
subject to clipping. A zero advantage removes this contribution; KL regularization may still
produce a gradient.

\Within{} normalizes rewards within each task--harness pair; \Cross{} pools harnesses for
each task. For rollout $i$ on task $x$ under harness $h$, let $r_{x,h,i}\in\{0,1\}$ be the
binary task-success reward. The corresponding advantages are
\begin{equation}
\begin{aligned}
A^{\Within{}}_{x,h,i}
&=\frac{r_{x,h,i}-\mu_{x,h}}{\sigma_{x,h}+\varepsilon}, \\
A^{\Cross{}}_{x,h,i}
&=\frac{r_{x,h,i}-\mu_x}{\sigma_x+\varepsilon}.
\end{aligned}
\label{eq:creditrules}
\end{equation}
Here $\mu$ and $\sigma$ denote the reward mean and standard deviation, computed per
task--harness pair for \Within{} and per task for \Cross{}. The constant $\varepsilon>0$
stabilizes the denominator. Both rules use identical trajectories and rewards, changing only the
normalization group. Pooling yields nonzero advantages when rewards differ across harnesses
but are constant within each harness. Section~\ref{sec:attribution} analyzes this change in
signal coverage and the harness information retained in the advantages.

\subsection{Controlled Training Setup}
\label{sec:training_setup}

Both RL arms start from the same Qwen3-8B SFT checkpoint and train by
offline replay of a fixed dataset with precomputed advantages. Trajectories are collected on
SWE-Gym \citep{pan2025swegym} through Aider, OpenHands, Qwen Code, and SWE-agent. We retain
$183$ tasks whose pooled rollouts contain both successful and unsuccessful attempts. The
resulting replay manifest contains $81{,}216$ training records derived from $5{,}543$ episodes;
an episode is one agent rollout on a task under a particular harness
(Appendix~\ref{app:training}).

We compare both RL variants with SFT to measure improvement from training, and use the
\Cross{}--\Within{} comparison to estimate the effect of reward grouping
(Equation~\ref{eq:creditrules}).
The primary arms share data, rewards, loss masks, optimizer settings, and one epoch of replay
($81{,}200$ optimizer steps). The pretrained Qwen3-8B model (Base) provides an additional
reference. Appendix~\ref{app:implementation} documents the training settings and
implementation checks.

Two auxiliary ablations examine whether \Cross{}'s performance is sensitive to the number of
trajectories with nonzero advantage and to differences in mean reward across harnesses.
\emph{Coverage-matched} sets some \Cross{} advantages to zero to match \Within{}'s number of
trajectories with nonzero advantage. \emph{Residualized} subtracts a per-harness mean-reward
baseline before pooled normalization.
The four RL arms---\Within{}, \Cross{}, Coverage-matched, and Residualized---all start from
SFT and share the replay data and update budget. Section~\ref{sec:signal_controls} details the
controls and evaluates their effects.

\subsection{Evaluation Protocol}
\label{sec:evaluation}

We evaluate each model variant under source and held-out harnesses to measure training gains
and their dependence on the execution interface. Evaluation uses $500$ SWE-bench
Verified tasks \citep{jimenez2024swebench,chowdhury2024verified}, disjoint from the training
tasks, under the four source harnesses and a minimal \weakreact{} harness excluded from
training. A sealed
oracle grades candidate patches using the same hidden tests under every harness; these tests
are separate from the feedback tests available to the agent. For an evaluation budget of
$k$ attempts per task, we report avg@$k$ by averaging binary success over graded attempts
within each task, then averaging the resulting task-level rates.

SFT-relative gains are reported separately under source and held-out harnesses to measure
improvement on the training interfaces and transfer to the unseen interface. Evaluating each
model variant across harnesses also places the training gains in the context of differences
between execution interfaces. The primary normalization comparison is the
\Cross{}--\Within{} difference on the held-out harness. We report mean paired differences
in task-level success rates with task-bootstrap confidence intervals ($95\%$ unless stated
otherwise). Infrastructure failures are excluded from scoring.
Appendix~\ref{app:pins} lists software versions; Appendix~\ref{app:evaluation} provides
complete results and statistical details.

\section{Performance and Portability}
\label{sec:performance}

We first quantify RL gains over SFT under the source harnesses and compare their scale with
performance variation across execution interfaces. We then evaluate transfer to an unseen
harness and sensitivity to reward normalization, followed by an analysis of action
distributions. All evaluations use tasks excluded from training.

\subsection{Training Gains and Harness Dependence}
\label{sec:harness}

Averaged over the four source harnesses, both normalization rules have higher solve rates
than SFT, with estimated gains of $+0.48$ percentage points (pp) for \Within{} and
$\PfSeenGain\pp$ for \Cross{}
(Figure~\ref{fig:spread}). The evaluation crosses six model variants---Base, SFT, \Within{},
\Cross{}, Coverage-matched, and Residualized---with the four source harnesses, using two
attempts per task. Both rules achieve their largest observed gain under OpenHands.
The \Cross{}--SFT estimate has a $99\%$ confidence interval (CI) of \PfSeenCIcorr{} after
Bonferroni correction for five comparisons, which includes zero
(Table~\ref{tab:contrastspf}).

\begin{paperfigure}[!b]
  \centering
  \includegraphics[width=0.8\linewidth]{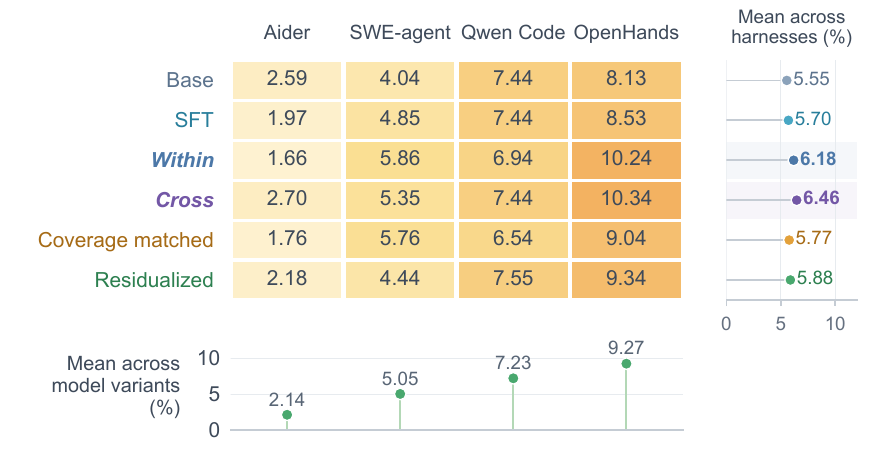}
  \caption{Source-harness performance on SWE-bench Verified at two attempts per task
  (\PfNevals{} attempted episodes). Rows are model variants and columns are harnesses;
  cells show avg@2 solve rates (\%), graded by the sealed oracle. Each bottom dot averages
  a harness's column over the six model variants; each right-hand dot averages a model's row
  over the four harnesses. The two sets of means use the same scale. Column means range from
  \PfColMin{}\% (Aider) to \PfColMax{}\% (OpenHands); row means range from
  \PfRowMin{}\% (Base) to \PfRowMax{}\% (\Cross{}). Harnesses are ordered by column mean.
  Table~\ref{tab:matrix} reports the corresponding task counts.}
  \label{fig:spread}
  \vspace{-4mm}
\end{paperfigure}

These estimated training gains are small relative to the performance variation across
evaluation harnesses. Averaging over model variants, the highest and lowest harness solve
rates differ by $7.13\pp$. Averaging over harnesses, the corresponding difference between
model variants is $0.91\pp$. This descriptive comparison shows substantially larger variation
across interfaces than across the tested training configurations, consistent with prior
model--harness evaluations
\citep{yao2026harnessbench,zhang2026stopcomparing,halharness2025}. We next test whether the
source-harness training gains extend to an unseen interface.

\subsection{Portability to the Held-Out Harness}
\label{sec:core}

\paragraph{Main results.}
On the held-out \weakreact{} harness, gains over SFT are small and statistically inconclusive
for both normalization rules. At eight attempts per task, the estimated gains are
$\PfWithinSftKeight\pp$ for \Within{} and $\PfHeldKeight\pp$ for \Cross{}, with both
$95\%$ CIs containing zero. The primary normalization comparison also yields a small
difference: \Cross{} exceeds \Within{} by $\PfCrossWithinKeight\pp$
($95\%$ CI: \PfCrossWithinKeightCI{}). These paired comparisons use different task subsets;
Table~\ref{tab:contrastspf} reports their selection criteria and confidence intervals.

\begin{paperfigure}[tb]
  \centering
  \includegraphics[width=0.85\linewidth]{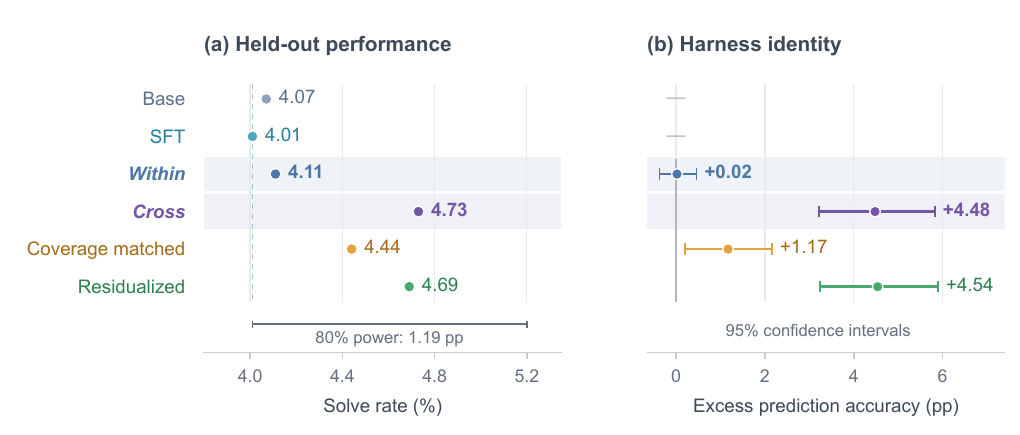}
  \caption{Held-out performance and harness identity in advantage estimates.
  \textbf{(a)} Solve rate on the held-out \weakreact{} interface at four attempts per task,
  over $499$ SWE-bench Verified tasks gradeable in every arm. The dashed line marks SFT.
  The ruler marks the smallest estimated solve-rate difference detectable at $80\%$ power
  among the six planned contrasts, which exceeds the spread of the six model variants.
  Tables~\ref{tab:heldout500} and~\ref{tab:contrastspf} report levels and paired contrasts.
  \textbf{(b)} Accuracy of predicting the generating harness from the scalar advantage on the
  training dataset, minus each arm's own shuffled-label accuracy, in percentage points with task-bootstrap
  $95\%$ intervals. Dashes indicate that the advantage diagnostic does not apply to Base
  or SFT. \Within{} shows no detectable excess; \Cross{} shows $\PfAdvCross\pp$.
  Section~\ref{sec:attribution} analyzes this diagnostic and the two auxiliary controls;
  Table~\ref{tab:advpred} reports the full results.}
  \label{fig:core}
  \vspace{-4mm}
\end{paperfigure}

The evaluation of all six model variants at four attempts per task also yields small
held-out differences (Figure~\ref{fig:core}(a)). Every planned pairwise comparison has a
$95\%$ CI containing zero (Table~\ref{tab:contrastspf}). Across both evaluation budgets,
the intervals remain compatible with modest gains or losses; these results therefore do
not establish equivalence between the training rules.

\paragraph{Sensitivity analyses.}
We assess whether the held-out comparison changes with the evaluation budget, training seed,
or a refresh of the replay data. Increasing attempts per task from four to eight improves
precision while leaving the conclusion of the primary comparison unchanged. On matched
task sets, the \Cross{}--\Within{} estimate moves closer to zero
(Table~\ref{tab:ksens}). Appendix~\ref{app:stats} reports the corresponding detection
thresholds and statistical details.

The direction of the \Cross{}--\Within{} difference varies across training seeds. We train
both variants with two additional seeds and evaluate all three pairs at four attempts
per task. Averaging the paired differences across seeds gives $\PfSeedCW\pp$
($95\%$ CI: \PfSeedCWCI{}). For each normalization rule, the difference between the highest
and lowest solve rates across seeds exceeds this mean \Cross{}--\Within{} difference.
Appendix~\ref{app:stats} gives the per-seed results and task-level breakdown.

To examine sensitivity to fixed replay data, we additionally evaluate a \Cross{} run that
refreshes its trajectories once during training. The run shares the SFT initialization and
total training budget of offline \Cross{}; its second half uses trajectories collected by
the partially trained policy. At four attempts per task, its held-out solve-rate gain over
offline \Cross{} is $\PfOnlineCross\pp$ ($95\%$ CI: \PfOnlineCrossCI{}), providing no
statistically significant improvement from this single data refresh. Appendix~\ref{app:crossed}
reports the full protocol and additional comparisons.

The held-out results leave the transfer of the observed source-harness gains unresolved
under either normalization rule, with no consistent advantage from pooling rewards.

\subsection{Action Distributions Across Harnesses}
\label{sec:behavior}

We next examine whether the harness dependence observed in solve rates also appears in
tool-use patterns. We compare action distributions in two ways: between harnesses for a
fixed model variant, and between SFT and \Cross{} under a fixed harness. A rule-based parser
labels each assistant turn by its primary action, such as search, inspection, editing,
testing, or submission. We compute the fraction of turns in each category within an episode
and average these shares across tasks. We then compare the distributions using
Jensen--Shannon divergence (JSD), measured in bits: zero indicates identical distributions,
and larger values indicate greater differences in action frequencies.

Action distributions differ substantially more across harnesses than between SFT and
\Cross{} under a fixed harness (Figure~\ref{fig:actions}). For each model variant, we compare
all six pairs among \weakreact{}, OpenHands, Qwen Code, and SWE-agent. The resulting JSD
values range from $\PfJsdAcrossMin$ to $\PfJsdAcrossMax$ bits, compared with
$\PfJsdWithinMin$ to $\PfJsdWithinMax$ bits for SFT versus \Cross{} under the same harness
(Table~\ref{tab:jsd}). Even the smallest divergence across harnesses is at least
\PfJsdRatio{} times the largest divergence between SFT and \Cross{}. Changes in individual
action shares are detected under \weakreact{} and OpenHands, while no category changes
significantly under Qwen Code or SWE-agent (Appendix~\ref{app:actions}).

Harness dependence therefore appears in both solve rates and aggregate action frequencies.
These distributions describe how SFT and \Cross{} use each interface; portability is assessed
by held-out task success (Section~\ref{sec:core}). The action analysis does not compare
\Within{} with \Cross{} or assess the quality of individual decisions.

\begin{paperfigure}[tb]
  \centering
  \includegraphics[width=0.8\linewidth]{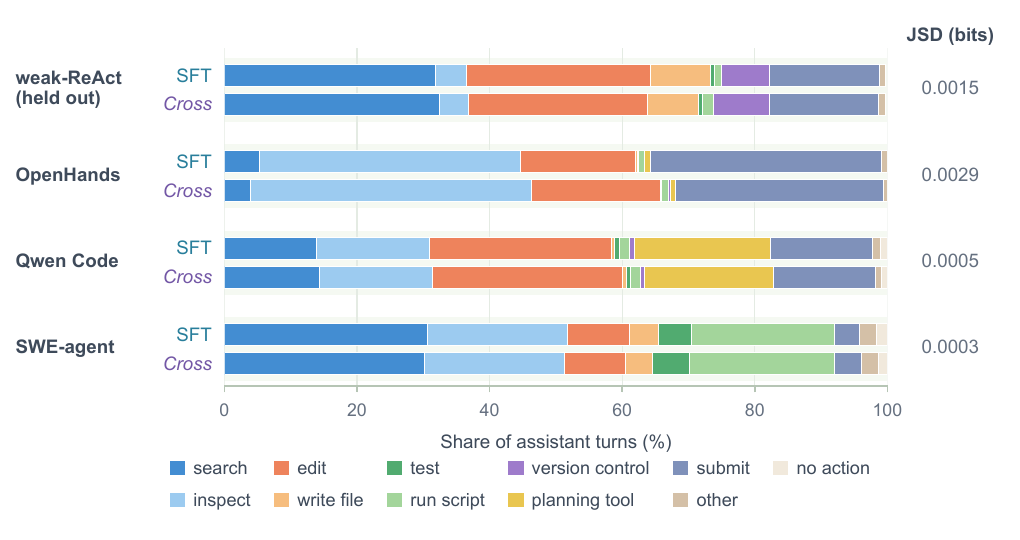}
  \caption{Action distributions of SFT and \Cross{} across evaluation harnesses. Bars show
  episode-level action shares averaged over $500$ SWE-bench Verified tasks, using the episodes
  behind Tables~\ref{tab:matrix} and~\ref{tab:heldout500}. Values at the right give the
  Jensen--Shannon divergence between SFT and \Cross{} in bits. Dominant action categories vary
  across harnesses, while the two model variants have similar distributions under each harness.
  Aider is omitted because its plain-chat transcripts contain no tool calls to label.
  Appendix~\ref{app:actions} describes the parser and its validation.}
  \vspace{-3.5mm}
  \label{fig:actions}
\end{paperfigure}

\section{Analysis of Advantage Estimates}
\label{sec:attribution}

The small held-out performance differences between \Within{} and \Cross{}
(Section~\ref{sec:core}) coexist with measurable changes in their training signals.
With the replay data fixed, pooling rewards assigns nonzero advantages to more trajectories
and preserves information about the generating harness. We quantify these changes, then
examine how the auxiliary controls modify them and affect held-out performance.

\subsection{Effects of Cross-Harness Normalization}
\label{sec:signal}

\paragraph{Nonzero advantages.}
Pooling increases the fraction of trajectories with nonzero advantage, which we refer to as
advantage coverage. Under \Within{}, $55.1\%$ of trajectories have zero advantage; under
\Cross{}, every trajectory has nonzero advantage (Table~\ref{tab:signal}). This difference
follows from the normalization groups: a task--harness group with identical binary rewards
assigns zero advantage to every trajectory, whereas every selected training task contains both successes
and failures when its trajectories are pooled across harnesses. Full coverage under
\Cross{} therefore depends on this task-selection criterion. Both rules replay all
trajectories; the grouping changes their contribution to the reward-dependent term in
Equation~\ref{eq:policy_objective}. Appendix~\ref{app:training} gives the group-level counts.

\paragraph{Harness identity.}
Pooling also retains differences in mean reward between harnesses. Within-harness
normalization centers each task--harness group separately, whereas pooled normalization
subtracts a single mean for the task. If a harness's mean reward on a task exceeds the pooled
mean, its trajectories have a positive mean advantage under \Cross{}.

We measure this association by predicting the generating harness from the scalar advantage
alone, with training and test tasks kept disjoint. We report \emph{excess accuracy}: the
classifier's test accuracy minus its baseline accuracy after shuffling harness labels within
tasks. Harness identity is predictable above this baseline for \Cross{}, with
$\PfAdvCross\pp$ excess accuracy ($95\%$ CI: \PfAdvCrossCI{}), but not detectably so for
\Within{}, whose excess is $\PfAdvWithin\pp$ (Figure~\ref{fig:core}(b)).
Appendix~\ref{app:e1} reports the full diagnostic protocol and results.

\subsection{Ablations of the Pooled Advantage}
\label{sec:signal_controls}

The two auxiliary controls test how held-out performance responds to changes in advantage
coverage and harness reward differences. Both modify \Cross{}'s precomputed advantages while
preserving the SFT initialization, replay data, and update budget. For each control, we
first examine its effect on the advantage diagnostics, then compare its held-out solve rate
with \Cross{} at four attempts per task.

\paragraph{Coverage matching.}
Coverage matching tests whether \Cross{}'s greater advantage coverage contributes to held-out
performance. It sets $55.1\%$ of \Cross{}'s advantages to zero, matching \Within{}'s nonzero
count while keeping all trajectories in training. The remaining advantages retain their
original pooled values.

This operation also reduces the classifier's excess accuracy to $\PfAdvCov\pp$
(Figure~\ref{fig:core}(b)), changing both coverage and harness predictability. The held-out
solve-rate difference, Coverage-matched minus \Cross{}, is $-0.28\pp$
($95\%$ CI: $[-1.27,+0.65]$; Table~\ref{tab:contrastspf}), with no statistically significant
performance change.

\paragraph{Reward residualization.}
Reward residualization tests whether subtracting differences in harness success rates changes
held-out performance. It subtracts a per-harness mean-reward baseline, estimated from other
tasks, before normalizing the adjusted rewards across harnesses for each task.
Appendix~\ref{app:objectives} gives the full definition in Equation~\ref{eq:resid}.

The resulting adjustment leaves harness predictability largely unchanged. The baseline is
estimated on the full collection pool, most of whose tasks failed under every harness,
whereas training retains tasks with mixed outcomes. It therefore adjusts for reward
differences in a different task population. Under Residualized, the classifier's
excess accuracy remains $\PfAdvResid\pp$, close to \Cross{}'s $\PfAdvCross\pp$
(Figure~\ref{fig:core}(b)). The held-out solve-rate difference, Residualized minus \Cross{},
is $\PfQsix\pp$ ($95\%$ CI: \PfQsixCI{}; Table~\ref{tab:contrastspf}).

The controls thus change the targeted advantage properties jointly or incompletely. They
characterize the sensitivity of the pooled advantages, but leave each property's contribution
to portability unresolved. Establishing that contribution requires an intervention that
separates the properties and verifies the intended change on the training distribution.

\section{Discussion and Conclusion}
\label{sec:discussion}
\label{sec:conclusion}

Our controlled study shows that multi-harness RL yields modest estimated gains over SFT on
the source harnesses under both \Within{} and \Cross{}, while performance differences across
evaluation harnesses remain substantially larger. The normalization comparison places these
gains in context: pooling changes how the same training experience contributes to policy
updates, but provides no consistent advantage on the held-out harness. Neither rule
significantly improves over SFT on that interface. Learning from multiple harnesses thus
leaves substantial dependence on the execution interface, even when the training data and
optimization settings are controlled.

These findings motivate treating policy optimization and the execution harness as joint
components of agent performance. Source-harness gains measure improvement under the
interfaces used to collect experience; their portability is a separate outcome that requires
evaluation under new interfaces. Our controlled design makes this distinction explicit while
isolating reward grouping from changes in the training data. Online training further couples
policy updates to subsequent experience collection, allowing the interaction between the
policy and its training harnesses to evolve. Extending our comparison to repeated data
collection and holding out different harnesses in turn would test how newly collected
experience affects transfer. The key direction is to identify training choices whose gains
persist across interfaces, alongside improvements in the harnesses through which policies act.

\section*{Limitations}
\label{sec:limitations}

Our study measures RL gains and the effect of reward normalization conditional on a fixed
set of multi-harness trajectories. The contribution of harness diversity itself requires
a matched comparison with single-harness training, which is outside this design.
In online training, policy updates change the trajectories subsequently collected, while
harness adaptation can change the tools and feedback available to the policy. The fixed-data
comparison does not characterize these interactions between learning, exploration, and
the execution interface.

The study's causal interpretation is limited to the normalization intervention. Each harness
combines tools, observations, and interaction protocols, so performance differences between
harnesses do not identify the contribution of any individual component. Likewise, the
advantage and action-distribution diagnostics characterize the training signal and observed
behavior without separating improvements in general problem-solving ability from adaptation
to a particular interface. The auxiliary controls leave this attribution unresolved because
they do not independently manipulate the relevant properties of the training signal.

Portability is measured by hidden-test success in repository-level code repair. This captures
benchmark correctness, while software maintainability and interaction efficiency remain
outside the evaluation. The findings come from a model configuration with low absolute
solve rates; their applicability to higher-performing agents and other interactive domains
remains untested.

\bibliographystyle{plainnat}
\bibliography{references}

\appendix
\clearpage
\begingroup
  \renewcommand{\contentsname}{Appendix Contents}
  \setcounter{tocdepth}{-1}
  \tableofcontents
\endgroup
\addtocontents{toc}{\protect\setcounter{tocdepth}{2}}
\medskip

\section{Training and Implementation Details}
\label{app:implementation}

This appendix documents the data and optimization settings behind the controlled
comparison, followed by the software versions and infrastructure checks used to keep
experimental conditions consistent.

\subsection{Training Data and Replay Diagnostics}
\label{app:training}

\paragraph{Data sources and units.}
All training trajectories come from SWE-Gym \citep{pan2025swegym}, and task-success
evaluation uses SWE-bench Verified. The training and evaluation task sets used in this study
are disjoint. The SFT corpus contains $8{,}841$ training records from successful trajectories
on $186$ tasks. The RL dataset contains $81{,}216$ training records derived from $5{,}543$
episodes on $183$ tasks. An episode is a complete agent rollout; the record count describes
the processed training data and should be distinguished from both the episode count and the
number of optimizer steps. Table~\ref{tab:tasksets} summarizes the task sets.

\paragraph{Task selection and advantage coverage.}
The $183$ RL tasks are selected from the $1{,}008$-task collection pool because their
trajectories contain both successes and failures when pooled across harnesses. For binary
rewards, this condition gives a positive reward standard deviation and nonzero advantages
under \Cross{}. A group in which every attempt succeeds or every attempt fails instead
assigns zero advantage to all trajectories \citep{yu2025dapo}.

Under \Within{}, the same trajectories form $698$ task--harness groups, of which $44.8\%$
contain mixed outcomes. The remaining groups have identical rewards and account for
$55.1\%$ of all trajectories. These two percentages use different denominators: the first
counts groups, while the second counts trajectories. Under \Cross{}, all $183$ task groups
have mixed outcomes, so every trajectory has nonzero advantage. Coverage matching then sets
$55.1\%$ of these pooled advantages to zero. Its reward-variance statistic remains $100\%$
because zeroing advantages does not change the rewards or the normalization groups.

\paragraph{Reward differences across source harnesses.}
On the selected RL tasks, the training-time solve rates are $6.9\%$ for Aider, $9.8\%$ for
SWE-agent, $20.9\%$ for Qwen Code, and $27.0\%$ for OpenHands. These rates describe the
selected training trajectories, not test-set performance. Their differences motivate the
per-harness reward baseline in the Residualized control
(Appendix~\ref{app:objectives}).

\paragraph{Offline replay.}
The four primary RL variants make one pass over a fixed dataset with precomputed advantages,
each completing $81{,}200$ logged optimizer steps. Consequently, reward variance and advantage
coverage describe the training data rather than changes in policy success during training.
The re-collected \Cross{} variant in Appendix~\ref{app:crossed} separately evaluates one
mid-training data refresh.

Table~\ref{tab:signal} reports the advantage and replay diagnostics. Mean importance ratios
range from $0.996$ to $0.999$, and the logged mean effective-sample-size shares are at least
$0.993$, indicating little concentration of the importance weights.

\begin{papertable}[htbp]
  \papertablecaption{tab:signal}{Replay diagnostics for the four RL variants. ``Nonzero var.'' is the
  percentage of normalization groups with nonzero reward variance; ``Zero adv.'' is the
  percentage of trajectories assigned zero advantage. RMS is the root mean square of
  advantages, and ESS is the logged mean effective-sample-size share of importance-weighted
  replay. All variants use the same data and number of optimizer steps.}
  \centering
  \small
  \begin{tabular}{lrrrrrr}
\toprule
Variant & Groups & Nonzero var. & Adv.\ RMS & Zero adv.\ (\%) & Steps & ESS \\
\midrule
\Within{} & 698 & 44.8\% & 0.670 & 55.1 & 81{,}200 & 0.996 \\
\Cross{} & 183 & 100.0\% & 1.000 & 0.0 & 81{,}200 & 0.995 \\
Coverage-matched & 183 & 100.0\% & 0.685 & 55.1 & 81{,}200 & 0.993 \\
Residualized & 183 & 100.0\% & 1.000 & 0.0 & 81{,}200 & 0.994 \\
\bottomrule
\end{tabular}
 \end{papertable}

\begin{papertable}[htbp]
  \papertablecaption{tab:tasksets}{Training and evaluation task sets. Training trajectories come from SWE-Gym;
  task-success evaluation uses SWE-bench Verified. The training and evaluation tasks used
  in this study are disjoint.}
  \centering
  \small
  \begin{tabular}{llp{0.52\linewidth}}
    \toprule
    Set & Size & Role \\
    \midrule
    Training pool & 1,008 tasks & SWE-Gym collection pool and source of the per-harness reward baselines \\
    SFT corpus & 186 tasks & Resolved trajectories from the four harnesses, $8{,}841$ records \\
    RL manifest & 183 tasks & Training tasks with both successes and failures in the pooled
    trajectories, yielding nonzero pooled advantages (Appendix~\ref{app:training}) \\
    \addlinespace[2pt]
    Verified eval pool & 500 & SWE-bench Verified; task-success evaluation \\
    \bottomrule
  \end{tabular}
\end{papertable}

\subsection{Training Objectives and Model Variants}
\label{app:objectives}

\paragraph{Reference models.}
Base is the pretrained Qwen3-8B model. SFT is obtained by supervised fine-tuning on successful
trajectories from the four source harnesses and initializes all RL variants. Comparisons
with SFT measure the additional gains from RL. SFT uses the same learning rate as RL,
$3\times10^{-7}$, and a matched compute budget; its hyperparameters are not separately tuned
for success-only distillation.

\paragraph{Shared RL configuration.}
\Within{}, \Cross{}, Coverage-matched, and Residualized use the same SFT initialization,
frozen multi-harness dataset, tokenizer, and chat template. Replay preserves the exact token
IDs sampled during trajectory collection, and observation tokens are masked from the loss.
All four variants use a learning rate of $3\times10^{-7}$, a KL coefficient of $0.01$, one
epoch, and a matched GPU-allocation budget, completing $81{,}200$ optimizer steps each
(Table~\ref{tab:signal}). The re-collected \Cross{} variant shares this setup but replaces
the data for the second half of training (Appendix~\ref{app:crossed}).

\paragraph{Advantage definitions.}
Let an episode be $e=(x,h,j)$, where $x$ denotes its task, $h$ its generating harness, and
$j$ its rollout index. Its binary task-success reward is $r_e\in\{0,1\}$. For a group $G$,
$\bar r_G$ and $\sigma_G$ denote the reward mean and standard deviation, and $\varepsilon$
is a numerical stability constant. \Within{} and \Cross{} apply the GRPO advantage
estimator \citep{shao2024deepseekmath} to different groups of the same trajectories.
Coverage matching modifies the resulting pooled advantages; residualization modifies
rewards before pooled normalization.

\paragraph{\Within{}.} For each task, trajectories from the same harness form a group,
$G_{\mathrm{w}}(x,h,j)=\{(x,h,j'):\forall j'\}$,
and the advantage is the group-standardized reward
\begin{equation}
A_e^{\Within{}} \;=\; \frac{r_e - \bar r_{G_{\mathrm{w}}(e)}}{\sigma_{G_{\mathrm{w}}(e)}+\varepsilon} .
\end{equation}
Each task--harness group is centered separately, so its mean advantage is zero.

\paragraph{\Cross{}.} Groups pool harnesses within a task,
$G_{\mathrm{c}}(x,h,j)=\{(x,h',j'):\forall h',j'\}$, and $A_e^{\Cross{}}$ is the same
group-standardized reward over $G_{\mathrm{c}}$. Pooled centering subtracts the task mean while
retaining differences in mean reward across harnesses within each task, as analyzed in
Section~\ref{sec:attribution}.

\paragraph{Coverage-matched.} This control tests sensitivity to the number of trajectories
with nonzero advantage. Starting from \Cross{}'s pooled advantages, it retains a subset of
nonzero values equal in size to \Within{}'s nonzero-advantage count and sets the others to
zero. All trajectories remain in replay, and the retained nonzero advantages keep their
original pooled values.

\paragraph{Residualized.} A per-harness baseline is subtracted from the reward before the pooled
standardization. The baseline for harness $h$ applied to task $x$ is the mean reward across
graded episodes from $h$ on other tasks, so task $x$'s own outcomes do not enter its baseline:
\begin{equation}
\begin{gathered}
b_h^{(-x)} \;=\; \operatorname{mean}\{\, r_{e'} : h(e')=h,\; x(e')\neq x \,\},
\paperlayout{\qquad}{\\}
z_e = r_e - b_{h(e)}^{(-x(e))}, \\
A_e^{\mathrm{Resid}} \;=\; \frac{z_e - \bar z_x}{\sigma^{z}_x+\varepsilon}.
\end{gathered}
\label{eq:resid}
\end{equation}
Here $x(e)$ and $h(e)$ denote episode $e$'s task and harness, respectively, and $z_e$ is its
reward after subtracting the baseline. The quantities $\bar z_x$ and $\sigma^z_x$ are the
mean and standard deviation of these residual rewards pooled across harnesses for task $x$.
The baseline is subtracted at unit coefficient, with no slope fitted. It is estimated over
the full graded pool of $1{,}008$ tasks, most of which failed under every harness, while
training uses the $183$ tasks with mixed pooled outcomes. The targeted offset is the
difference in mean reward across harnesses on these selected training tasks
(Appendix~\ref{app:training}). Because the collection pool and selected training subset
have different reward distributions, this baseline need not remove the differences among
harnesses in the training data. Section~\ref{sec:signal_controls} evaluates the resulting
advantages using harness predictability and held-out performance.

\subsection{Harness and Runtime Versions}
\label{app:pins}

Table~\ref{tab:pins} records the installed versions used throughout the study. These
include the four source harnesses
\citep{gauthier2024aider,wang2025openhands,qwencode2025,yang2024sweagent}, the held-out
\weakreact{} implementation based on ReAct \citep{yao2023react}, and the serving, training,
and grading runtimes \citep{kwon2023vllm,jimenez2024swebench}. Holding these versions fixed
keeps software changes from entering comparisons between model variants.

\begin{papertable}[htbp]
  \papertablecaption{tab:pins}{Harness and Runtime Versions (frozen for the whole campaign).}
  \centering
  \small
  \begin{tabular}{lll}
    \toprule
    Component & Package & Version \\
    \midrule
    Aider & \texttt{aider-chat} & 0.86.2 \\
    OpenHands & \texttt{openhands-sdk} (V1 API) & 1.0.0a6 \\
     & \texttt{openhands-ai} (co-installed) & 0.62.0 \\
    Qwen Code & \texttt{qwen-code} & 0.19.3 \\
    SWE-agent & \texttt{sweagent} / \texttt{swe-rex} & 1.1.0 / 1.4.0 \\

    \weakreact{} & this work & single-loop ReAct implementation \\
    \midrule
    Serving & \texttt{vllm} & 0.8.5.post1 \\
    Training & \texttt{torch} / \texttt{transformers} / \texttt{verl} & 2.6.0 / 4.51.3 / 0.4.1 \\
    Grading oracle & \texttt{swebench} / \texttt{udocker} & 4.1.0 / 1.3.17 \\
    \bottomrule
  \end{tabular}
\end{papertable}

\subsection{Infrastructure Validation}
\label{app:audit}

Infrastructure checks preserve the independence of rollouts and distinguish unsuccessful
task attempts from failures to execute or grade an attempt. During data collection and
evaluation, the following issues were detected, corrected, and checked again before the
corresponding results entered analysis.

\begin{enumerate}
  \item \textbf{Shared evaluation containers.} Shared container stores allowed edits from one
  experimental condition to appear in another condition's evaluation. Creating a private
  container store for each rollout job prevented this contamination.
  \item \textbf{Model-server readiness.} A model-registration race produced HTTP 404 responses
  that were treated as permanent task skips, dropping approximately $20\%$ of episodes in
  affected collection batches. Readiness probes were updated to check model registration;
  affected episodes were quarantined and re-collected.
  \item \textbf{Text-decoding failures during grading.} Strict UTF-8 decoding caused grader
  crashes, which reduced the number of valid verdicts. Replacing invalid byte sequences during
  decoding allowed grading to proceed. Infrastructure errors remained distinct from valid
  task-success and task-failure verdicts and were excluded from score denominators.
  \item \textbf{Corrupted Git repositories.} Missing HEAD files, damaged references, and
  truncated pack indexes prevented task repositories from building. The last failure could
  pass a \texttt{rev-parse} check. In-container checkout verification and rebuilds addressed
  these cases.
  \item \textbf{Incomplete image caches.} A partially populated image cache truncated every
  collection job at the same $25$ tasks. A cache-integrity check before collection prevented
  this systematic omission of tasks.
\end{enumerate}

\section{Supplementary Results and Statistical Analysis}
\label{app:evaluation}

We first report the complete performance results and auxiliary evaluations, then detail
the scoring, task subsets, and uncertainty estimates used to interpret these comparisons.

\subsection{Performance Results}
\label{app:crossed}

This subsection provides the complete performance tables for Section~\ref{sec:performance} and
additional details on the data-refresh and second-model experiments. The six primary model
variants are Base, SFT, \Within{}, \Cross{}, Coverage-matched, and Residualized.
All four RL variants start from the same SFT model and replay the same multi-harness data;
their advantage estimates differ as specified in Appendix~\ref{app:objectives}.
Comparisons with SFT measure the gains from RL, while \Cross{}--\Within{} comparisons
measure the effect of changing the normalization groups.

\paragraph{Reading the tables.}
Tables~\ref{tab:matrix} and~\ref{tab:heldout500} report absolute solve rates in percent.
Table~\ref{tab:contrastspf} reports paired differences in percentage points (pp), with
positive values favoring the first named model. Each source-harness column uses the tasks
with valid scores for all six variants. Held-out absolute rates use the common tasks for
all variants evaluated at that attempt budget, whereas paired contrasts use the task
subsets specified in Appendix~\ref{app:stats}. Consequently, a paired difference need not
equal the difference between two displayed absolute rates. Table~\ref{tab:ksens} compares
four and eight attempts per task on matched task sets.

\begin{papertable}[htbp]
  \papertablecaption{tab:matrix}{Source-harness solve rates (\%) on SWE-bench Verified, measured by avg@2
  using hidden-test oracle verdicts. Each column uses the tasks scoreable for all six model
  variants. Row means average the four harness columns; column means average the six model
  variants. The last row gives the number of tasks in each column.}
  \centering \small
  \begin{tabular}{lrrrrr}
\toprule
Model variant & Aider & OpenHands & Qwen Code & SWE-agent & Row mean \\
\midrule
Base (pretrained) & 2.59 & 8.13 & 7.44 & 4.04 & \textbf{5.55} \\
SFT warm start & 1.97 & 8.53 & 7.44 & 4.85 & \textbf{5.70} \\
\Within{} & 1.66 & 10.24 & 6.94 & 5.86 & \textbf{6.18} \\
\Cross{} & 2.70 & 10.34 & 7.44 & 5.35 & \textbf{6.46} \\
Coverage-matched & 1.76 & 9.04 & 6.54 & 5.76 & \textbf{5.77} \\
Residualized & 2.18 & 9.34 & 7.55 & 4.44 & \textbf{5.88} \\
\midrule
Column mean & \textbf{2.14} & \textbf{9.27} & \textbf{7.23} & \textbf{5.05} & \\
Tasks in column & 482 & 498 & 497 & 495 & \\
\bottomrule
\end{tabular}
 \end{papertable}

\begin{papertable}[htbp]
  \papertablecaption{tab:heldout500}{Held-out solve rates (\%) under \weakreact{}, the harness excluded from training.
  The avg@4 column uses $499$ tasks common to all six variants; avg@8 uses $500$ tasks common
  to SFT, \Within{}, and \Cross{}. The seed column reports avg@4 on $499$ tasks common to
  all six seed-specific \Within{} and \Cross{} models. Dashes indicate settings not evaluated.}
  \centering \small
    \begin{tabular}{lrrl}
\toprule
Model variant & avg@4 (\%) & avg@8 (\%) & avg@4 by seed 17 / 29 / 41 (\%) \\
\midrule
Base (pretrained) & 4.07 & --- & --- \\
SFT warm start & 4.01 & 4.38 & --- \\
\Within{} & 4.11 & 4.28 & 4.11 / 4.26 / 4.56 \\
\Cross{} & 4.73 & 4.38 & 4.73 / 4.31 / 4.36 \\
Coverage-matched & 4.44 & --- & --- \\
Residualized & 4.69 & --- & --- \\
\bottomrule
\end{tabular}
 \end{papertable}

\begin{papertable}[htbp]
  \papertablecaption{tab:contrastspf}{Paired solve-rate differences in percentage points, computed as the first named
  model minus the second. Intervals are $95\%$ task-bootstrap CIs unless labeled $99\%$.
  ``Strict pairs'' requires eight graded attempts per task for both models in the comparison;
  the resulting task subsets can differ across rows. Appendix~\ref{app:stats} details the
  task selection, multiplicity correction, and aggregation across training seeds.}
  \centering \small \setlength{\tabcolsep}{4pt}
  \begin{tabular}{llrcl}
\toprule
Evaluation & Contrast & $\Delta$pp & CI & Tasks / notes \\
\midrule
Source, four harnesses & \Cross{} $-$ SFT & +0.77 & [+0.03, +1.52] & 500, avg@2 \\
\quad corrected, five comparisons & & & [-0.21, +1.75] & 99\% CI \\
\midrule
Held-out, $k{=}4$ & SFT $-$ Base & -0.07 & [-0.92, +0.80] & 500 \\
\quad & \Within{} $-$ SFT & +0.10 & [-0.80, +0.95] & 500 \\
\quad & \Cross{} $-$ SFT & +0.72 & [-0.25, +1.72] & 500 \\
\quad & \Cross{} $-$ \Within{} & +0.62 & [-0.33, +1.62] & 500 \\
\quad & Coverage-matched $-$ \Cross{} & -0.28 & [-1.27, +0.65] & 499 \\
\quad & Residualized $-$ \Cross{} & -0.03 & [-1.03, +0.95] & 500 \\
\midrule
Held-out, $k{=}4$, three seeds & \Cross{} $-$ \Within{} & +0.16 & [-0.41, +0.72] & 499 \\
\midrule
Held-out, $k{=}4$, re-collected & Re-collected $-$ \Cross{} & +0.13 & [-0.85, +1.10] & 500 \\
\quad & Re-collected $-$ \Within{} & +0.75 & [-0.10, +1.65] & 500 \\
\quad & Re-collected $-$ SFT & +0.85 & [-0.15, +1.85] & 500 \\
\midrule
Held-out, $k{=}8$ & \Cross{} $-$ SFT & +0.13 & [-0.65, +0.92] & strict pairs \\
\quad & \Within{} $-$ SFT & -0.03 & [-0.64, +0.59] & strict pairs \\
\quad & \Cross{} $-$ \Within{} & +0.25 & [-0.48, +1.02] & strict pairs \\
\bottomrule
\end{tabular}
 \end{papertable}

\begin{papertable}[htbp]
  \papertablecaption{tab:ksens}{Sensitivity to the evaluation budget. Solve-rate differences are in percentage
  points and use matched task sets for the four- versus eight-attempt comparison. The last
  row gives the range of estimated minimum detectable effects at $80\%$ power across
  contrasts; Appendix~\ref{app:stats} defines this measure.}
  \centering \small
  \begin{tabular}{lrr}
\toprule
Held-out contrast & $k{=}4$ & $k{=}8$ \\
\midrule
\Cross{} $-$ \Within{} & +0.58 & +0.25 \\
\Within{} $-$ SFT & +0.16 & -0.03 \\
\Cross{} $-$ SFT & --- & +0.13 \\
\midrule
Minimum detectable effect & 1.19--1.42 & 0.88--1.14 \\
\bottomrule
\end{tabular}
 \end{papertable}

\paragraph{A single data refresh during RL.}
The re-collected \Cross{} variant tests whether updating the rollout data changes held-out
performance. It uses the same SFT initialization and total update budget as offline \Cross{},
but replays the original data only during the first half of training. The intermediate policy
then collects new trajectories on the $183$ training tasks through the four source harnesses.
Of these tasks, $131$ still contain both successful and unsuccessful attempts and supply
the second half of training. At its first update on the refreshed data, both the importance
ratio and effective-sample-size share are $1.000$, consistent with rollouts collected by the
intermediate policy.

On the $500$ tasks scoreable for this variant and offline \Cross{}, \Within{}, and SFT,
its held-out avg@4 is $4.85\%$. Its differences from offline \Cross{} and \Within{} are
$\PfOnlineCross\pp$ ($95\%$ CI: \PfOnlineCrossCI{}) and $\PfOnlineWithin\pp$
($95\%$ CI: \PfOnlineWithinCI{}), respectively. Neither difference is statistically
significant (Table~\ref{tab:contrastspf}). This experiment evaluates one data refresh for
\Cross{}; it does not compare repeated on-policy training under the two normalization rules.

\paragraph{A second pretrained model.}
Table~\ref{tab:crossmodel} evaluates harness dependence with Seed-Coder-8B and Qwen3-8B.
The upper block compares the two pretrained models under the same harness. The lower block
compares Aider and SWE-agent with each model's own \weakreact{} score, so these harness
comparisons do not require the two models to have equal baseline performance. Evaluation uses
four attempts per task under \weakreact{} and two under Aider and SWE-agent. Each row uses
its own paired task subset, which explains small differences from the Qwen3-8B entries in
Table~\ref{tab:matrix}.

OpenHands and Qwen Code produced no patches with Seed-Coder-8B: all $1{,}000$ episodes
under each ended in a harness error. The same error occurred in $61$ of $1{,}000$ OpenHands
episodes and $287$ of $999$ Qwen Code episodes with Qwen3-8B. These Seed-Coder-8B settings
are recorded as unmeasurable and omitted from the table. Its context limit is $32{,}768$
tokens, compared with $40{,}960$ for Qwen3-8B; the $32$ context overflows among its $2{,}000$
\weakreact{} episodes are counted as unsuccessful attempts.

\begin{papertable}[htbp]
  \papertablecaption{tab:crossmodel}{Performance under a second pretrained model family on SWE-bench Verified.
  The upper block reports solve rates (\%) and Seed-Coder-8B minus Qwen3-8B differences (pp).
  The lower block compares each harness with \weakreact{} within the same model (pp).
  Each row uses the $n$ tasks scoreable in both compared settings, with $95\%$
  task-bootstrap confidence intervals.}
  \centering \small
  \begin{tabular}{lrrrlr}
\toprule
Harness & Seed-Coder-8B & Qwen3-8B & $\Delta$pp & 95\% CI & $n$ \\
\midrule
weak-ReAct (minimal) & 5.00 & 4.07 & +0.93 & [-0.22, +2.13] & 500 \\
Aider & 0.62 & 2.59 & -1.97 & [-3.32, -0.83] & 482 \\
SWE-agent & 0.60 & 4.03 & -3.43 & [-4.94, -2.12] & 496 \\
\midrule
\multicolumn{6}{l}{\emph{Change from weak-ReAct within the same model}} \\
\quad Seed-Coder-8B, Aider & \multicolumn{2}{r}{} & -4.41 & [-5.86, -3.06] & 499 \\
\quad Seed-Coder-8B, SWE-agent & \multicolumn{2}{r}{} & -4.40 & [-5.85, -3.05] & 500 \\
\quad Qwen3-8B, Aider & \multicolumn{2}{r}{} & -1.42 & [-3.08, +0.24] & 482 \\
\quad Qwen3-8B, SWE-agent & \multicolumn{2}{r}{} & -0.07 & [-1.63, +1.51] & 496 \\
\bottomrule
\end{tabular}
 \end{papertable}

\subsection{Statistical Protocol}
\label{app:stats}

\paragraph{Scoring and paired estimation.}
All reported task-success scores use a hidden-test oracle, which re-executes each candidate
patch in a per-instance container. Its tests are disjoint from the feedback tests visible to
the model. An initial validation found $2.58\%$ disagreement between inline feedback and
oracle verdicts, exceeding the $1\%$ tolerance; all disagreements were optimistic inline
verdicts. We therefore use oracle verdicts throughout.

For each model--harness setting, avg@$k$ averages each task's success rate over its graded
attempts, then averages across tasks. A paired contrast averages the difference between two
models' task-level success rates over the eligible tasks. We compute two-sided $95\%$
percentile confidence intervals using $10{,}000$ bootstrap resamples of tasks with fixed
random seeds \citep{efron1994bootstrap,agarwal2021precipice}. Resampling tasks preserves
the dependence among repeated attempts on the same task \citep{cameron2015cluster}.
Infrastructure failures have no task-success verdict and are excluded from score denominators.

\paragraph{Coverage and task subsets.}
Every evaluation starts from the same $500$-task roster. A model--harness setting enters
analysis only if at least $95\%$ of attempted episodes receive valid task-success verdicts;
the analysis stops if this requirement is not met. The source-harness evaluation attempted
\PfNevals{} episodes and achieved \PfCoverage{}\% verdict coverage. Held-out coverage is
$98.0$--$99.7\%$ at $k{=}4$ and $98.6$--$99.3\%$ at $k{=}8$.

Absolute rates and paired contrasts use different task intersections. In
Table~\ref{tab:matrix}, each source-harness column includes tasks scoreable for all six
model variants, yielding $482$, $498$, $497$, and $495$ tasks for Aider, OpenHands,
Qwen Code, and SWE-agent, respectively. Held-out absolute rates in
Table~\ref{tab:heldout500} use $499$ tasks common to all six variants at $k{=}4$, and $500$
tasks common to SFT, \Within{}, and \Cross{} at $k{=}8$. Each task's rate uses its available
graded attempts. The $k{=}4$ paired contrasts use the task counts in
Table~\ref{tab:contrastspf}; they agree with differences between the displayed absolute
rates to within $0.01\pp$.

The $k{=}8$ contrasts apply a stricter completeness rule. We first retain the $486$ tasks
with at least seven graded attempts for each of SFT, \Within{}, and \Cross{}. Each paired
comparison then uses only tasks with all eight attempts graded for both models. The three
contrasts therefore use potentially different subsets and need not be additive. The
data-refresh comparisons use $500$ tasks scoreable for the re-collected variant and all
three offline comparators. Table~\ref{tab:ksens} uses matched task sets to compare attempt
budgets, which explains why its $k{=}4$ estimates differ slightly from the primary
$k{=}4$ contrasts.

\paragraph{Multiple comparisons.}
The six primary held-out $k{=}4$ contrasts and their order were fixed before the checkpoints
were produced. They were not externally preregistered \citep{vanmiltenburg2021prereg}.
For the source-harness \Cross{}--SFT contrast, Table~\ref{tab:contrastspf} reports both the
$95\%$ interval and a Bonferroni-corrected $99\%$ interval accounting for five model
comparisons. The latter includes zero.

\paragraph{Rollout and training-seed variation.}
Identical evaluation reruns disagree on approximately $6\%$ of per-task outcomes because
of continuous-batching nondeterminism. This is a task-level disagreement rate, not an
uncertainty interval for the aggregate solve rate. We use repeated attempts and paired,
task-level inference to account for evaluation variability
\citep{bouthillier2021accounting,madaan2024variance,miller2024errorbars}. Because benchmark
composition can affect model rankings \citep{dehghani2021benchmark}, all comparisons start
from the same task roster.

To examine training variability, we repeat \Within{} and \Cross{} with seeds 29 and 41
in addition to seed 17, evaluating each model with four attempts per task. On the $499$
tasks common to all runs, the \Cross{}--\Within{} differences are $+0.62$, $+0.05$, and
$-0.20\pp$ for seeds 17, 29, and 41. Averaging paired task-level differences across these
seeds gives $\PfSeedCW\pp$ ($95\%$ CI: \PfSeedCWCI{}). This interval resamples tasks
while retaining the three observed seeds. After averaging over seeds, \Cross{} has higher
task-level success on $42$ tasks, \Within{} on $38$, and the two tie on $419$. The
maximum-minus-minimum solve-rate ranges across seeds are $\PfSeedRangeW\pp$ for \Within{}
and $\PfSeedRangeC\pp$ for \Cross{} (Table~\ref{tab:heldout500}).

\paragraph{Evaluation budget and statistical power.}
On matched task sets, increasing attempts from four to eight changes the
\Cross{}--\Within{} difference from $\PfCrossWithinKfour\pp$ to $\PfCrossWithinKeight\pp$,
and the \Within{}--SFT difference from $+0.16\pp$ to $\PfWithinSftKeight\pp$
(Table~\ref{tab:ksens}). We estimate MDE$_{80}$, the true solve-rate difference required for
an $80\%$ chance of detecting a nonzero effect, from interval widths under a normal
approximation \citep{card2020power,miller2024errorbars}. Across contrasts, these estimates
range from \PfMdeKfour$\pp$ at $k{=}4$ to \PfMdeKeight$\pp$ at $k{=}8$;
the three-seed comparison has an estimated MDE$_{80}$ of
about $0.8\pp$. These values describe the sensitivity of the evaluation. A nonsignificant
contrast does not establish equivalence between the training rules.

\section{Advantage and Behavioral Diagnostics}
\label{app:diagnostics}

The following analyses characterize the training signal and the resulting action
distributions, complementing the task-success comparisons in Appendix~\ref{app:evaluation}.

\subsection{Harness Identity in Advantage Estimates}
\label{app:e1}

This diagnostic measures how much information about the generating harness is retained in
scalar advantages. It analyzes the fixed training weights used by each RL variant; policy
performance is evaluated separately in Appendix~\ref{app:crossed}.

\paragraph{Prediction protocol.}
We predict each trajectory's generating harness from its scalar advantage alone. Tasks are
hashed into five folds, and all trajectories from the same task remain in one fold. The
classifier is trained on four folds and evaluated on the remaining fold, producing
out-of-fold (OOF) predictions with disjoint training and test tasks.

For each RL variant, we repeat the procedure after shuffling harness labels within tasks,
using the same folds. This gives an empirical baseline for that variant.
\emph{Excess accuracy} is OOF accuracy minus shuffled-label accuracy, in percentage points.
We compute its $95\%$ confidence interval with $10{,}000$ task-bootstrap resamples.
Table~\ref{tab:advpred} reports both accuracies and their difference, so the baseline used
for each variant is explicit.

\paragraph{Association with harness success rates.}
As a complementary diagnostic, we regress each variant's scalar advantages on the
training-time success rate of the trajectory's generating harness. The predictor uses the
same raw-reward success rates for every variant; only the advantages change. The resulting
slope measures association between advantage and harness success rate, while the classifier
measures recoverable harness identity. Table~\ref{tab:advpred} reports both diagnostics.

\begin{papertable}[htbp]
  \papertablecaption{tab:advpred}{Harness prediction from a scalar advantage on the fixed training dataset
  ($183$ tasks, $5{,}543$ episodes). OOF accuracy is evaluated on held-out task folds;
  excess accuracy subtracts each variant's shuffled-label baseline. Intervals apply to
  excess accuracy. The final column reports the regression slope described above.}
  \centering
  \small
  \resizebox{\linewidth}{!}{\begin{tabular}{lrrrlr}
\toprule
Variant & OOF acc.\ (\%) & Shuffled baseline (\%) & Excess (pp) & 95\% CI & Slope \\
\midrule
\Within{} & 26.41 & 26.39 & +0.02 & [-0.38, +0.46] & -0.00 \\
Coverage-matched & 27.46 & 26.29 & +1.17 & [+0.20, +2.16] & +0.94 \\
\Cross{} & 30.63 & 26.15 & +4.48 & [+3.22, +5.83] & +2.38 \\
Residualized & 30.65 & 26.12 & +4.54 & [+3.24, +5.90] & +2.05 \\
\bottomrule
\end{tabular}
 }
\end{papertable}

\subsection{Action-Distribution Analysis}
\label{app:actions}

\paragraph{Action labeling and aggregation.}
We analyze the evaluation transcripts underlying Tables~\ref{tab:matrix}
and~\ref{tab:heldout500}. A rule-based parser assigns each assistant turn one primary action
label, using the priority order submit, edit, write, test, run, search, inspect, version
control, planning tool, and other. We compute action shares within each episode and average
them across tasks. Aider is excluded because its plain-chat transcripts lack the tool calls
or shell commands used by this parser.

Compound commands receive the highest-priority applicable label. For example, a turn that
edits a file and then runs tests counts as an edit. As a sensitivity check, allowing multiple
labels per turn raises the held-out testing share from $0.6\%$ for both models to $3.2\%$
for SFT and $2.9\%$ for \Cross{}. These values reflect different labeling conventions for
the same transcripts.

\paragraph{Parser validation.}
A manual spot check found $11$ mislabeled turns out of $160$, including $8$ under SWE-agent.
The analysis retains three tasks whose framework containers wrote empty files into the
working tree; excluding them changes no action share by more than $0.2\pp$.

\paragraph{Changes in action shares.}
On the held-out \weakreact{} harness, \Cross{} allocates a smaller share of actions to
whole-file writing than SFT ($\PfActWrite\pp$; $95\%$ CI: \PfActWriteCI{}), and larger
shares to script execution ($\PfActRun\pp$; $95\%$ CI: \PfActRunCI{}) and version control
($\PfActVcs\pp$; $95\%$ CI: \PfActVcsCI{}). These differences subtract SFT's mean
episode-level action share from that of \Cross{}. No statistically significant change
is detected in search, inspection, editing, or testing. Under OpenHands, inspection and
editing shares increase; under Qwen Code and SWE-agent, no action category has a
statistically significant change.

\paragraph{Comparing distributions.}
Jensen--Shannon divergence summarizes differences in action-category frequencies: zero
indicates identical distributions, and larger values indicate greater differences. We compare
SFT with \Cross{} under each harness, then compare the six pairs of the four parsed
harnesses while holding the model fixed (Table~\ref{tab:jsd}). These aggregate action
frequencies describe behavior under each interface; they do not measure whether individual
actions contribute to solving a task.

\begin{papertable}[H]
  \papertablecaption{tab:jsd}{Jensen--Shannon divergence (JSD) between action distributions, in bits.
  The upper block compares SFT and \Cross{} under the same harness on $500$ paired tasks,
  with $95\%$ task-bootstrap intervals. The lower block gives the minimum and maximum
  divergence across the six harness pairs for each model variant; these ranges are not
  confidence intervals.}
  \centering
  \small
  \begin{tabular}{llrl}
\toprule
Comparison & Harness or model variant & JSD & 95\% CI \\
\midrule
SFT vs \Cross{}, same harness & \weakreact{} & 0.0015 & [0.0009, 0.0033] \\
 & OpenHands & 0.0029 & [0.0016, 0.0077] \\
 & Qwen Code & 0.0005 & [0.0004, 0.0027] \\
 & SWE-agent & 0.0003 & [0.0003, 0.0029] \\
\midrule
Same model variant, harness pairs & SFT & 0.177--0.344 & \\
 & \Cross{} & 0.174--0.350 & \\
\bottomrule
\end{tabular}
\end{papertable}

\section{Context for Reported Multi-Harness Gains}
\label{app:relatedtab}

Existing systems report gains from training agents through execution harnesses under
different comparison protocols. The examples below distinguish gains under a training
harness from transfer to an excluded harness, and changes to a complete training recipe
from changes to advantage normalization.

\paragraph{OpenForge RL.}
On ClawEval, a personal-assistant and tool-use benchmark, the ZeroClaw-only SFT+RL model
improves pass@1 over the base model by $3.3\pp$ under unseen OpenClaw and $4.6\pp$ under
unseen Codex. Training with ZeroClaw, OpenClaw, and Codex raises the base-relative gains to
$9.5\pp$ and $20.3\pp$ under OpenClaw and Codex, respectively; both are source harnesses
for this three-harness model. The comparison changes harness exposure during both SFT and
RL \citep{openforge2026}.

\paragraph{Orchard.}
Orchard's controlled SFT comparison trains separate models using trajectories from
mini-swe-agent or OpenHands, then evaluates both on SWE-bench Verified under both harnesses.
Solve rates are $53.5$--$57.9\%$ when the training and evaluation harnesses match, versus
$19.0$--$28.0\%$ when they differ. Its full SFT model also reaches $45.0\%$ under unseen
Kimi-CLI; this is an absolute solve rate for the combined SFT recipe \citep{orchard2026}.

\paragraph{Polar.}
Polar trains Qwen3.5-4B with GRPO separately under four coding harnesses and evaluates each
resulting model under its training harness. On SWE-bench Verified, pass@1 improves over the
base model by $22.6\pp$ under Codex, $4.8\pp$ under Claude Code, $0.6\pp$ under Qwen Code,
and $6.2\pp$ under Pi. These results show that the size of the observed RL gain varies
substantially across training--evaluation harness pairs \citep{polar2026}.

Our study complements these system-level results by holding the multi-harness trajectories
and optimization settings fixed, comparing two reward-normalization rules, and evaluating
both under source and held-out harnesses. This design relates the measured RL gains to
harness-dependent performance while isolating the normalization choice.

\section{Reproducibility and Research Transparency}
\label{app:transparency}

\subsection{Reproducibility and Release}
\label{app:repro}

Appendix~\ref{app:implementation} documents the training data, objectives, software
versions, and infrastructure checks. Appendix~\ref{app:stats} specifies grading and
statistical inference; Appendix~\ref{app:diagnostics} describes the advantage and
action-distribution diagnostics.

Continuous-batching nondeterminism can produce different rollouts when an evaluation is
repeated with the same model and settings. The reported comparisons therefore use repeated
attempts and paired, task-level statistics.

The planned release will include the evaluation harness, grading oracle, analysis code,
recorded results, and analysis plan. These materials are not public at submission time.

\subsection{Broader Impacts and Ethics}
\label{app:broader}

Harness-dependent performance affects how agent benchmarks inform model selection,
deployment, and safety evaluation. In our source-harness matrix, mean solve rates averaged
over the six model variants range from $2.14\%$ to $9.27\%$ across harnesses. A score obtained
under one execution interface may therefore give an incomplete picture of performance under
a different deployment interface. Safety evaluations can likewise miss behaviors enabled
by tools or feedback absent from the evaluated harness.

Reporting the harness, task set, grader, and number of attempts makes these comparisons
easier to interpret. Separating gains measured under source harnesses from gains under a
held-out harness further clarifies the evidence for transfer. The additional evaluation
compute supports more informative deployment and benchmarking decisions.

\subsection{AI Assistance}
\label{app:aiuse}

We used generative AI tools, including Claude (Anthropic) and ChatGPT (OpenAI), to refine
hypotheses, review methods and analysis plans, check interpretations, edit prose, generate
and refactor code, design figures, and support literature searches. ChatGPT was also used
for review-style critique of drafts.

The authors reviewed all AI-assisted analysis code and checked AI-suggested citations
against the cited sources. Reported numbers come from analysis scripts applied to experiment
result files, not from generated text. Generative AI was not used to fabricate, select, or
alter experimental data or verdicts. The authors take responsibility for the final text,
claims, code, and artifacts.

\end{document}